\documentclass[letterpaper, 10 pt, conference]{ieeeconf}  %
\usepackage{graphics} %
\usepackage{subcaption}
\usepackage{epsfig} %
\usepackage{hyperref}
\usepackage{times} %
\usepackage{amsmath} %
\usepackage{amssymb}  %
\usepackage{graphicx} %
\usepackage[dvipsnames]{xcolor}
\usepackage{xcolor}
\usepackage{booktabs}
\usepackage{cite}
\usepackage{tikz}

\newcommand{\textcirclednum}[1]{\textcircled{\raisebox{-0.6pt}{\scalebox{0.8}{#1}}}}

\definecolor{ouryellow}{RGB}{255,204,0}

\usepackage{caption}
\newcommand{\setcaptionformat}{
  \captionsetup[figure]{justification=justified, singlelinecheck=false, font=footnotesize}
}

\hypersetup{
  colorlinks=true,
  linkcolor=blue,
  filecolor=magenta,      
  urlcolor=cyan,
  citecolor=blue,
}

\renewcommand{\ref}[1]{\autoref{#1}}

\usepackage{afterpage}
\makeatletter
\newcommand{\setcaptype}[1]{\def\@captype{#1}}
\makeatother
\newsavebox{\tempbox}

\IEEEoverridecommandlockouts                              %

\title{\LARGE \bf
QueSTMaps: Queryable Semantic Topological Maps for 3D Scene Understanding}

\author{
  Yash Mehan$^{1}$\textsuperscript{*}, 
  Kumaraditya Gupta$^{1}$\textsuperscript{*}, 
  Rohit Jayanti$^{1}$\textsuperscript{*}, 
  Anirudh Govil$^{1}$, 
  Sourav Garg$^{2}$, and 
  Madhava Krishna$^{1}$%
  \thanks{The authors acknowledge the support provided by MeitY,
Govt. of India, under the project ``Capacity building for
human resource development in Unmanned Aircraft System
(Drone and related Technology)”.}
  \thanks{*Denotes authors with equal contribution}%
  \thanks{
    $^{1}$Robotics Research Center, IIIT-Hyderabad, India
  }%
  \thanks{$^{2}$The University of Adelaide, Australia}
  }%

\begin{document}

\maketitle
\thispagestyle{empty}
\pagestyle{empty}

\setcaptionformat
\vspace{-10pt} %

\savebox{\tempbox}{\begin{minipage}{\textwidth} %
\setcaptype{figure}%
    \centering
    \includegraphics[width=1.0\linewidth]{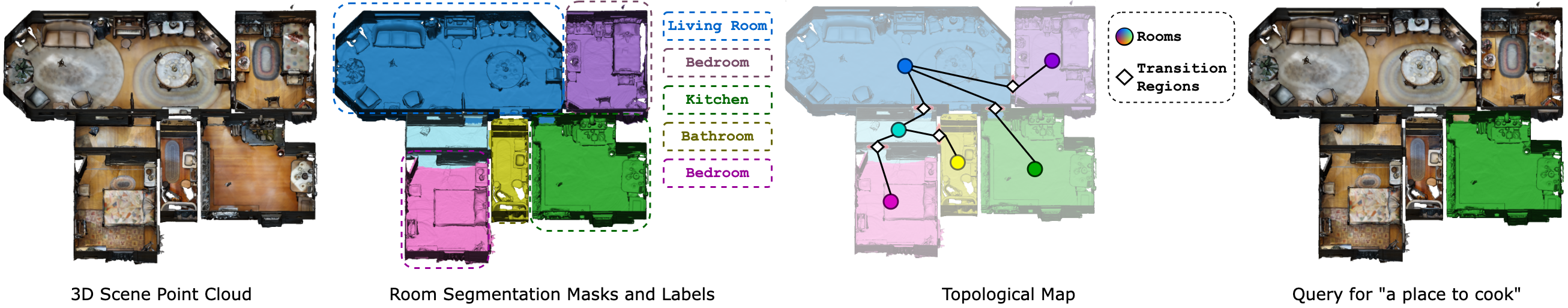}
    \captionsetup{justification=justified, singlelinecheck=false, font=footnotesize}
    \caption{\textbf{Topology Extraction and Room Labeling for Indoor Scene 3D Point clouds.} Given a 3D scene point cloud we (a) predict room and transition regions, (b) generate room-label aligned CLIP embeddings to assign room labels, and (c) build a topological map that supports room-level natural language queries, e.g., the query \texttt{place to cook} locates the \texttt{kitchen}.}
    \label{fig:teaser-wide}
    \vspace{-10pt} %
\end{minipage}}

\begin{figure}[t]
\rlap{\usebox\tempbox}
\end{figure}
\afterpage{\begin{figure}[t]%
\rule{0pt}{\dimexpr \ht\tempbox+\dp\tempbox}
\end{figure}}

\begin{abstract}

Robotic tasks such as planning and navigation require a hierarchical semantic understanding of a scene, which could include multiple floors and rooms. Current methods primarily focus on object segmentation for 3D scene understanding. However, such methods struggle to segment out topological regions like ``kitchen'' in the scene. In this work, we introduce a two-step pipeline to solve this problem. First, we extract a topological map, i.e., floorplan of the indoor scene using a novel multi-channel occupancy representation. Then, we generate CLIP-aligned features and semantic labels for every room instance based on the objects it contains using a self-attention transformer. Our language-topology alignment supports natural language querying, e.g., a ``place to cook'' locates the ``kitchen''. We outperform the current state-of-the-art on room segmentation by $\sim$20\% and room classification by $\sim$12\%. Our detailed qualitative analysis and ablation studies provide insights into the problem of joint structural and semantic 3D scene understanding. Project Page: \href{https://quest-maps.github.io/}{quest-maps.github.io}

\end{abstract}

\section{Introduction}

Understanding both the \textit{structure} and the \textit{semantics} of 3D indoor environments is essential for mapping and navigation in real world~\cite{han2021semantic, crespo2020semantic,zimmerman2022long}. The structure, i.e., the layout or floorplan, of an indoor environment defines its organisation into different rooms or spaces. Each room has a semantic label associated with it, e.g., \texttt{kitchen}, which can be inferred from the semantic labels of the objects inside it. Thus, we can generate a topological understanding of such 3D scenes using connections \textit{both} within and across the rooms, that is, objects-to-room and room-to-room. Such a topological understanding coupled with a semantic understanding helps plan paths across several rooms, when given a free-form natural language query.

Very few existing works address the twin problem of room segmentation and classification, with most focusing on segmentation alone. Floorplan extraction methods attempting to understand \textit{only the structure} of the scene~\cite{chen2022heat}, ignore the rich semantics essential for high-level robotic tasks~\cite{muravyev2023evaluation, garg2020semantics}. On the other hand, dense 3D semantic segmentation methods~\cite{peng2023openscene, seichter2021efficient} have focused \textit{solely on object-level} semantic segmentation, without considering the room-level structure of indoor scenes~\cite{takmaz2023openmask3d}.

We propose a first-of-its-kind method that generates a natural language-queryable room-level topological and semantic representation of a 3D indoor scene, by segmenting then understanding. We propose a novel multi-channel occupancy representation of 3D point clouds for room segmentation. These channels help in disambiguating the transition regions (e.g., doors) and rooms. We generate room-level features for each room segment using the CLIP~\cite{clip} embeddings of objects confined in them. We align the room features with natural language descriptions of the room using a self-attention transformer. As a result, our method supports free-form natural language queries to detect relevant rooms. For example, the query \texttt{a place to cook} locates the \texttt{kitchen}.
We benchmark our method on synthetic datasets like Structured3D~\cite{zheng2020structured3d} and challenging real-world datasets like Matterport3D~\cite{Matterport3D}. We achieve $\sim$20\% and $\sim$12\% performance gain over the previous SoTA methods on floorplan extraction~\cite{yue2023connecting, chen2022heat} and room labeling~\cite{chen2023leveraging} respectively. Our contributions are:
\begin{enumerate}
    \item  a novel pipeline to extract \textit{both} the \textit{topology} (rooms and transition regions) and \textit{semantic room labels} from 3D indoor scenes;
    \item a novel \textit{multi-channel occupancy representation} of  point clouds that improves room and transition region detection;
    \item an \textit{attention-based} method to generate \textit{representations for rooms based on the CLIP embeddings of objects in them}, enabling natural language room-level queries; 
    \item an extended version of the Matterport3D dataset~\cite{Matterport3D}, with \textit{manual annotations of transition regions} in the scenes.
\end{enumerate}

 \vspace{-10pt} %
\section{Related Work}

\subsection{Floorplan Reconstruction Methods}

Floorplan reconstruction organises raw sensor data like point clouds, density maps, or RGB-D images into a structured representation of rooms. Classical methods leverage techniques like plane-fitting~\cite{mura2014automatic} or model the problem as an energy minimisation problem~\cite{ambrucs2017automatic}. The heuristic-based detection of such methods struggles with noisy and incomplete real world 3D data. Deep learning based methods tackle this concern~\cite{he2018mask, chen2019floorsp, liu2018floornet}, where some methods~\cite{carion2020endtoend, NEURIPS2023_987bed99} such as HEAT~\cite{chen2022heat} use transformers to generate point, edge, or room level proposals to generate a floorplan. These methods do not predict semantics of the indoor environment (i.e., no room labels like \texttt{bedroom,} \texttt{kitchen}, etc.) and are thus limited in their downstream utility~\cite{crespo2020semantic}.

\subsection{Semantic Segmentation of Floorplans}
This includes assigning a semantic label to every detected room, and transition regions like doorways and windows. FloorNet~\cite{liu2018floornet} utilises a CNN-DNN architecture and also identifies salient objects in the scene. However, its object set is very limited. RoomFormer~\cite{yue2023connecting}, the current state-of-the-art, models the problem as a polygon fitting task by regressing to vertices and edges. Although such a formulation suits synthetic scenes~\cite{zheng2020structured3d} with straight edges and regular angles, it struggles on real-world scenes~\cite{Matterport3D} which invariably contain noise, ill defined edges, and curved walls. These methods also struggle to segment rooms with cluttered objects. Since these methods do not perform dense semantic segmentation, they do not have open vocabularies and cannot be queried in natural language.

\subsection{3D Semantic Segmentation}
3D semantic segmentation methods assign a label to 3D pixels or voxels in the scene~\cite{couprie:hal-00805105}. Methods like OpenScene~\cite{peng2023openscene, seichter2021efficient} use per-pixel image features extracted from posed RGB-D images of a scene and obtain a point-wise task-agnostic scene representation but struggle in differentiating instances of objects. 3D instance segmentation methods~\cite{takmaz2023openmask3d,han2020occuseg, xie2021unseen} distinguish multiple objects belonging to the same semantic category, predicting individual masks for each instance, but do not build structural associations among objects. Thus, they cannot label entire regions like kitchens, bedrooms etc.

\subsection{3D Scene Understanding and Scene Graphs}

For a robot to understand a scene it must combine place and object labels in a hierarchical manner to create a hybrid map~\cite{garg2020semantics}.~\cite{pham2016geometrically} uses an unsupervised geometry-based approach to segment 3D point clouds into objects and meaningful scene structures but its segmentations lack semantics.~\cite{sunderhauf2016place}  can recognize new semantic classes online, but does not leverage rich 3D object geometries and is limited to semantic segmentation in 2D space.
To unite structure with semantics~\cite{pham2018scenecut,bavle2022situational,Armeni_2019_ICCV,liao2020tsm} seek to build 3D scene graphs where nodes represent objects and edges encode inter-object relationships. To perform downstream tasks we must be able to query these graphs conveniently. ConceptGraphs and others~\cite{conceptgraphs, garg2023robohop,chen2023open,strader2023indoor} use LLMs to accomplish this due to their open-vocabulary and reasoning abilities.
However, all these methods struggle to segment rooms from unstructured data, and cannot retrieve all objects from a queried room.

\begin{figure*}[t]
    \centering
    \includegraphics[width=\textwidth]{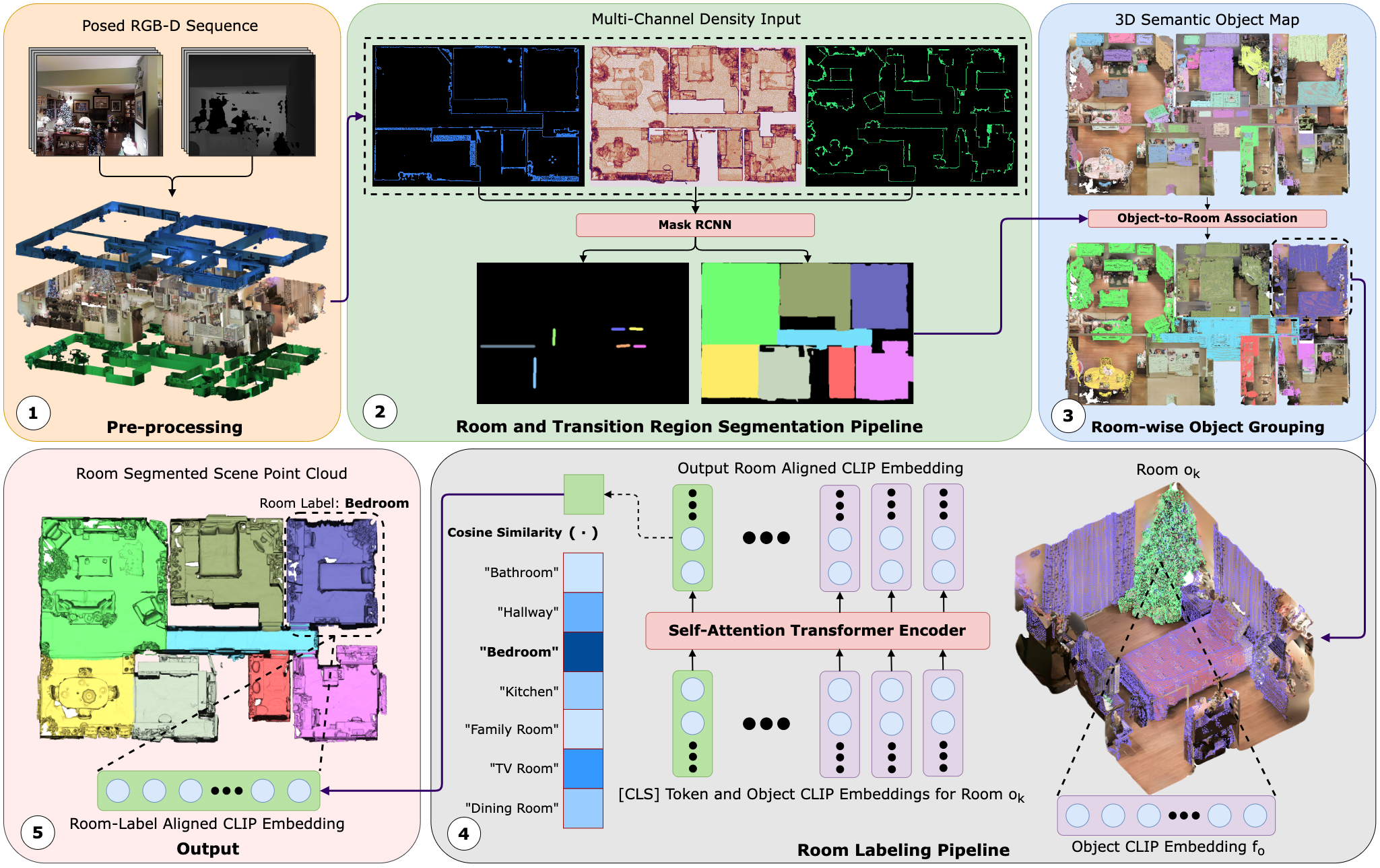}
    \caption{\textbf{Method Overview.} From a 3D scene point cloud reconstructed from a posed RGB-D sequence, our method builds an indoor scene topology map, extracts room and transition region segments, and generates room-label aligned CLIP embeddings and room labels. We propose a novel multi-channel input representation and use an instance segmentation network to predict room masks and detect transition regions. Object level point clouds and CLIP embeddings are generated using an instance mapping pipeline similar to ConceptGraphs~\cite{conceptgraphs} and are associated with their rooms from the previous step. A Transformer network takes these object level CLIP embeddings to generate a room label and an aligned CLIP embedding, useful for tasks such as languge guided robot navigation with a room-level understanding.}
    \label{fig:pipeline-wide}
    \vspace{-10pt} %
\end{figure*}

\section{Method}
Our pipeline is illustrated in Fig.~\ref{fig:pipeline-wide}. Our work focuses on an offline setting. Given a set of posed RGB-D images captured in an indoor scene we construct a point cloud and a 3D semantic object instance map using ConceptGraphs~\cite{conceptgraphs}. We use this as our foundation to predict room-level segmentation masks and room labels. We formulate the problem of room segmentation as an instance mask prediction problem, predicting masks for both rooms and transition regions. We perform instance segmentation using a Mask R-CNN architecture, that uses our novel multi-channel occupancy representation to predict room segments. We use a self-attention transformer encoder to produce label-aligned features for each room instance, using the CLIP features~\cite{clip} of the objects inside it. This allows us to query the scene in natural language at a room topology level.

\subsection{Room and Transition Region Segmentation}\label{subsec:method-room-seg-tr-det}

\subsubsection{Multi-Channel Occupancy Representation}\label{subsubsec:input-channel}
We propose a novel multi-channel input representation for 3D point clouds. Prior works~\cite{yue2023connecting,chen2019floorsp,chen2022heat} relied solely on \textit{top-down} density maps as a consolidated 2D representation of the complete 3D point cloud. There are several drawbacks to this representation. Firstly, in sparse reconstruction methods, density of a point cloud of an indoor scene depends highly on the quality of the scanning process; the density in a specific region can vary considerably depending on the trajectory and the amount of time spent by the sensor scanning that specific region. Secondly, under relaxed assumptions about the wall thickness of the environment, it is hard to disambiguate key regions in a top-down density map, e.g. walls vs closed doors.
Our key insight is to introduce two additional input channels: two sliced occupancy maps built at floor and ceiling heights, respectively. That is, given a pointcloud $\mathcal{P} = \{p_1, p_2, \hdots, p_n\}$ consisting of $n$ unordered 3D points, $p_i \in \mathbb{R}^3, i=1, \hdots, n$ represented using Euclidian co-ordinates, we introduce a 3-channel input representation $\{D, O_{ceiling}, O_{floor}\}$. These are illustrated in Part {\textcirclednum{1}} and {\textcirclednum{2}} of Fig.~\ref{fig:pipeline-wide} (blue for ceiling and green for floor). While the standard technique employed by~\cite{yue2023connecting, chen2019floorsp} is used to obtain the top-down density map $D$, to generate the occupancy slices $\{O_{ceiling}, O_{floor}\}$ we construct a grid of the same resolution as the top-down density map and build occupancy maps of filtered points in a specified range of height values as follows:

 \vspace{-10pt} %
\begin{equation}
O_{ij} = \mathbb{I}\left(\sum_{p \in \mathcal{P}} \mathbb{I}(\beta_1 h \leq p_z \leq \beta_2 h) \geq 0\right)
\end{equation}

where $O_{ij}$ is the occupancy state of the grid cell, and $\beta_1$ and $\beta_2$ define the extent of the slice. For the ceiling slice, we choose $\beta_1 = 0.7$ and $\beta_2 = 0.9$, and for the floor slice, we choose $\beta_1 = 0.1$ and $\beta_2 = 0.3$. $p$ represents a point in the point cloud, $p_z$ is the z-coordinate of the point, $h$ is the height of the point cloud calculated as the difference between the ceiling and the floor heights, and $\mathbb{I}(\cdot)$ is the indicator function.

As illustrated in Fig.~\ref{fig:pipeline-wide}, these additional input channels clearly delineate open doorways as a difference between the floor and ceiling occupancy maps, simplifying the problem of doorway/transition region detection, and consequently room segmentation.

\subsubsection{Room-Mask Prediction and Transition Region Detection}
We formulate room instance mask prediction and transition region detection as a 3-class (2 classes + background) instance segmentation task. The input for this segmentation task is our three-channel 2D array consisting of the density map and the top and bottom occupancy maps, as detailed in Sec.~\ref{subsubsec:input-channel}.
The total loss in Mask R-CNN is given by:
\begin{equation}
\mathcal{L} = \mathcal{L}_{c} + \mathcal{L}_{\text{BBR}} + \mathcal{L}_{m}
\end{equation}
where \(\mathcal{L}_{c}\), \(\mathcal{L}_{\text{BBR}}\), and \(\mathcal{L}_{m}\) represent the classification (\(c\)), bounding box regression (\text{BBR}), and mask (\(m\)) losses, respectively. 

\textit{Classification loss} (\(\mathcal{L}_{c}\)) measures the discrepancy between the predicted and ground truth room and transition region class probabilities. \textit{Bounding box regression loss} (\(\mathcal{L}_{\text{BBR}}\)) penalizes the difference between predicted 2D room and transition region bounding box and ground truth bounding box coordinates of the respective class. 
\textit{Mask segmentation loss} (\(\mathcal{L}_{m}\)) evaluates the dissimilarity between predicted 2D room/transition masks and their corresponding ground truth binary masks.

\subsection{Room Labeling}\label{subsec:method-room-labeling}

We introduce a novel approach to assign room labels to the room masks generated in Sec.~\ref{subsec:method-room-seg-tr-det}. To achieve this, we combine these room segments with their contained objects, where we obtain object instances using a 3D semantic instance mapping pipeline~\cite{conceptgraphs}. Using the semantic information for the objects in a room, we use our model to obtain room embeddings aligned with the CLIP features of room labels. 

\subsubsection{3D Object Instance Map}
We use the semantic object map creation methodology described in ConceptGraphs~\cite{conceptgraphs} to obtain a dense object instance map, as detailed in this sub-subsection for the sake of completion. This map is derived from a series of posed RGB-D images, collectively represented as \(I = \{I_1, I_2, \ldots, I_n\}; I_t = \langle I_{t}^{rgb}, I_{t}^{depth}, \theta_t\rangle\). For some \(1 \leq t \leq n\). 
It lifts the outputs of 2D foundation models into 3D by aggregating observations associated across multiple views. This results in distinct point clouds and features for each object. Segment Anything Model (SAM)~\cite{Kirillov_2023_ICCV} is used to obtain class-agnostic instance masks within each image, denoted as \(\{m_{t,i}\}_{i=1}^M = \text{Seg}(I_{t}^{rgb})\). The corresponding mask features \(f_{t,i} = \text{Embed}(I_{t}^{rgb}, m_{t,i})\) are computed using CLIP~\cite{clip} by cropping the masked image. Each object within the scene, \(o_j\), is thus defined by a tuple \(o_j = \langle p_{o_j}, f_{o_j} \rangle\), where \(p_{o_j}\) represents the object's point cloud and \(f_{o_j}\) represents the associated CLIP embedding for the object. This point cloud \(p_{o_j}\) fuses the constituent point clouds \(p_i\) through multi-view projection of its 2D masks \(m_{t,i}\) into 3D using depth \(I_{t}^{depth}\) and pose \(\theta_t\). In doing so, CLIP features \(f_{t,i}\) are averaged across views to obtain a single object CLIP feature vector \(f_{o_j} \in \mathbb{R}^{D}, D=1024\). This feature encapsulates the semantic attributes of the object.

\subsubsection{Object-to-Room Association}
After obtaining the room segmentation masks \(R = \{r_1, r_2, \ldots, r_n\}\) as detailed in Sec.~\ref{subsec:method-room-seg-tr-det}, we allocate object instances to specific rooms. We apply a transformation to the segmentation masks to convert from 2D image space back to 3D world co-ordinates. This transformation is facilitated by the offset and tile-size metadata we save during the point cloud pre-processing step detailed in subsection~\ref{subsubsec:input-channel}

We employ a KD Tree construction for each room mask, to accurately relate discrete object instances to their corresponding room masks. incorporating all \((x, y)\) coordinates as defined in the world frame. For each object identified within the scene, we compute the centroid of its point cloud. This centroid serves as a query point against the constructed KD Trees, with the closest match dictating the room assignment for the respective object.

\subsubsection{Assigning Room Labels}
For each room \(r_k\), we have a subset of objects \(\{o^{r_k}_1, o^{r_k}_2, \ldots, o^{r_k}_u\}\). The CLIP features for these objects are encoded as a sequence of tokens, serving as input to a self-attending transformer encoder. We prepend a CLS (classification) token for room segment \(r_k\), to the sequence of object feature vectors, represented as \(S_k = [CLS; f_{o^{r_k}_1}; f_{o^{r_k}_2}; \ldots]\). The purpose of this token is to aggregate information across the sequence, with the attention weights determining the extent of each object's contribution. The CLS token encapsulates the room's semantic characteristics as \(E_k = T_{enc}(S_k)\) where \(E_k\) is the encoded representation.

During the training phase, we compute the cosine similarity between the CLS token \(e_{CLS} = E_k[0], e_{CLS} \in \mathbb{R}^D\) and the CLIP text embeddings of various room phrases, like \texttt{bedroom}, \(T = \{t_1, t_2, \ldots, t_v\}, t_v \in \mathbb{R}^D\). 

We aim to maximize the similarity between the CLS token embedding and the ground truth room label's text embedding using a contrastive loss function, similar to the Normalized Temperature-scaled Cross Entropy Loss. This refines the model's ability to output a room label aligned embedding \(e_{CLS}\) , which is in the CLIP embedding space.
\begin{equation}
\mathcal{L}(e_{CLS}, t_{pos}) = -\log\frac{\exp(\text{sim}(e_{CLS}, t_{pos}) / \tau)}{\sum_{j} \exp(\text{sim}(e_{CLS}, t_j) / \tau)}
\label{eq:nt_xent_loss}
\end{equation}
During inference, the room label is determined by identifying the room type phrase embedding that has the highest cosine similarity with the CLS token output embedding. We assign this inferred label to the 3D region. 
\begin{equation}
    \text{label} = \underset{t \in T}{\arg\max} \; \text{sim}(e_{CLS}, t)
\end{equation}
Additionally, the output is designated as the room embedding for the associated 3D region. This facilitates a targeted search for specific room types using natural language queries, since the room embedding is aligned with CLIP. 

\subsection{Training Details}
The Mask R-CNN model used for transition region and room mask segmentation uses a ResNet50 architecture as its backbone, utilizing the layer outputs 1, 2, 3, and 4 of the backbone. The model is configured with an anchor generator for the Region Proposal Network (RPN), a box RoI pooler, and a mask RoI pooler. We initialize the anchor generator with sizes set to (32,), (64,), (128,), and (256,) pixels, and the aspect ratios are configured to 0.5, 1.0, and 2.0 for each size. To reduce detection redundancy, we run Non-Maximum Suppression (NMS) with a threshold of 0.5. AdamW was used as the optimizer with a learning rate \(\alpha = 2 \times 10^{-4}\).

Our self-attention transformer model for room labeling incorporates an embedding size of 1024, with 8 attention heads, and 8 layers. During training, we set the dropout at a rate of 0.2 to prevent overfitting and the temperature \(\tau\) for the loss function to 0.5. Optimization is performed using AdamW, with a learning rate of \(1 \times 10^{-5}\) for 400 epochs.

\section{Experiments and Results}
Sec.~\ref{subsec:expt-tr-and-room-seg} discusses datasets, metrics, ablations, results, and comparisons of the room segmentation and transition region detection step. Sec.~\ref{subsec:expt-room-labeling} discusses labeling rooms and finding room labels for each room mask. In Sec.~\ref{subsec:expt-complete-pipeline} we evaluate the complete pipeline and benchmark it. We also show qualitative comparisons to highlight our method's ability to understand the topology of 3D scenes.

\subsection{Room and Transition Region Segmentation} \label{subsec:expt-tr-and-room-seg}
This part of the topological segmentation pipeline, utilises a MaskRCNN ~\cite{he2018mask} to segment a scene into instance-separated rooms as well as instance-separated doors, i.e. transitions between the rooms. 

\textbf{Datasets.} We evaluate our method on two datasets-Structured3D and Matterport3D~\cite{zheng2020structured3d, Matterport3D}. \textbf{Structured3D} comprises panoramic Manhattan and non-Manhattan floorplan layouts with semantically rich room annotations with synthetic scenes with minimal clutter. Notably, the rooms within this dataset are geometrically defined as polygons, a feature that has facilitated the efficacy of previous works ~\cite{yue2023connecting,chen2022heat} based on polygon fitting techniques. We adhere to the official split of 3000 training, 250 validation, and 250 test samples. \textbf{Matterport3D} provides RGBD scans and navigable meshes of real-world indoor scenes with object and room type annotations. It is used extensively for robotic tasks like navigation and 3D scene understanding. It offers very detailed scenes with cluttered rooms. The major challenges associated with it are the smaller number of scenes (90), class imbalance in room types, and noisy meshes. 
Although Matterport3D is a popular dataset for 3D indoor scene understanding, no other work discusses topological segmentation of scenes for this dataset.
Since Matterport3D does not provide annotations for transition regions, we manually annotated all doorways in the dataset. We share this extension to the dataset with the community. 

\textbf{Evaluation Metrics.} Since we model the room segmentation task as a region-based instance segmentation problem, we use the Average Precision (AP) @ 0.5 IoU threshold as it considers both the accuracy of positive predictions and the ability to capture all instances of a target object. Since the baselines we compare against~\cite{yue2023connecting,chen2022heat} output an ordered set of vertex coordinates, we consider their predicted mask as a polygon constructed by connecting such vertices. \cite{chen2022heat} does not segment transition regions, hence the cells have been left blank in Table \ref{tab:tr_room_table}.

\begin{table}[htbp]
    \centering
    \caption{\textbf{Quantitative Results for Room Segmentation.}}
    \begin{tabular}{@{}lcccc@{}}
        \toprule
        Method & \multicolumn{2}{c}{Structured3D} & \multicolumn{2}{c}{Matterport3D} \\
        \cmidrule(lr){2-3} \cmidrule(lr){4-5}
        & Transitions & Rooms  & Transitions & Rooms  \\
         & (AP) & (AP)& (AP)& (AP) \\
        \midrule
        HEAT~\cite{chen2022heat} & - & 94.24 &  - & 30.78 \\
        RoomFormer~\cite{yue2023connecting} & 76.94 & 96.93 & 24.85 & 64.42 \\
        \midrule
        Ours (Single Channel) & 85.60 & 94.90 & 50.47 & 84.83 \\
        Ours (Multi Channel) & \textbf{87.99} & \textbf{98.86} & \textbf{60.92} & \textbf{88.47} \\
        \bottomrule
    \end{tabular}
    \label{tab:tr_room_table}
\end{table}

\begin{figure}[htbp]
\centerline{\includegraphics[width=\columnwidth]{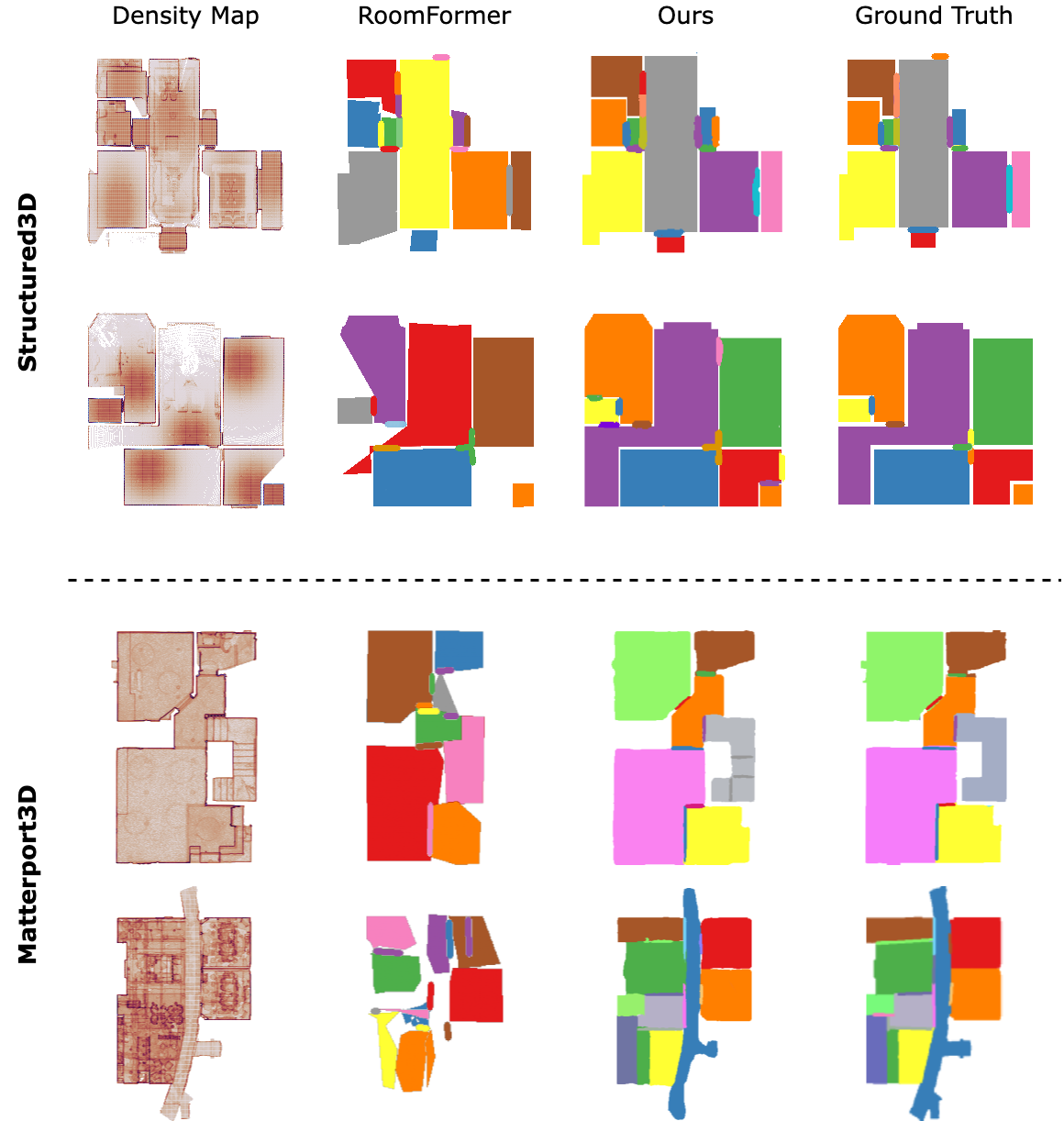}}
\caption{\textbf{Qualitative comparisons on room segmentation task.} Our proposed method is able to handle adverse cases such as regions of low density (introduced by the scanning strategy) in the Stuctured3D dataset, as well as curved walls and clutter as seen in the Matterport3D dataset. Colors are only representative of the instance separation.}
\label{fig}
\vspace{-10pt} %
\end{figure}

\textbf{Baselines and Results.}
We primarily compare our model against the state-of-the-art RoomFormer~\cite{yue2023connecting} and HEAT~\cite{chen2022heat}. Both are transformer-based methods that fit polygons and output an ordered set of vertices. \cite{chen2022heat} first detects room corners, and then predicts edges, but no doorways. Results are summarized in Table \ref{tab:tr_room_table}, where we outperform the average precision of the state-of-the-art RoomFormer by 1.9\% on Structured3D and by 20\% on Matterport3D. In addition to rooms, RoomFormer also detects transition regions i.e. doorways. Our method registers a 10\% higher average precision than the SoTA RoomFormer on doorway detection. To demonstrate the effectiveness of our multi-channel representation (top and bottom occupancy maps), we construct a baseline that learns only on a single-channel density map. As in Table~\ref{tab:tr_room_table}, our ablations exhibit an improvement in both doorway detection and room segmentation tasks.

\subsection{Room Labeling} \label{subsec:expt-room-labeling}

\textbf{Datasets.} In this section we conduct our experiments on the Matterport 3D (MP3D) dataset. Other datasets like Structured3D do not provide posed RGB-D sequences of the scene. So, MP3D serves as the foundation for our model evaluation and facilitates a comprehensive comparison with existing approaches.

\begin{figure*}[htbp]
    \centering
    \includegraphics[width=\textwidth]{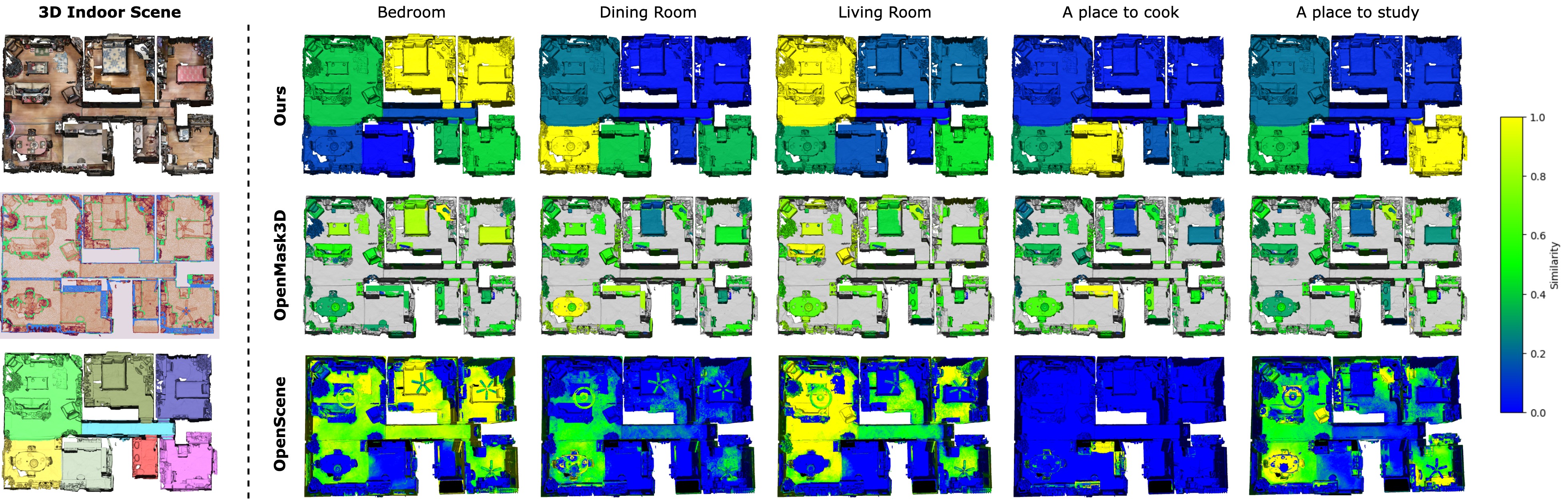}
    \caption{\textbf{Qualitative results on Matterport3D.} \underline{Left}: Original 3D Scene Point cloud, our multi-channel input representation, and the predicted room segmentation masks for reference. \underline{Right}: Qualitative comparisons for various room label queries. Here sections of the point cloud are colored based on the similarity of the corresponding query embedding. Color progression from blue to green to yellow signify \textcolor{blue}{low} to \textcolor{green}{medium} to \textcolor{ouryellow}{high} similarity scores.}
    \label{fig:qualitative-wide}
    \vspace{-10pt} %
\end{figure*}

\textbf{Evaluation Metrics.} We evaluate performance using two metrics: the weighted average F1 score and Mean Average Precision (mAP), across 23 room label types present in the dataset. Metrics for learning-based approaches are calculated after training the model on the MP3D dataset's train split (80\%), and evaluated on the test split (20\%).

\textbf{Baseline Methods.} To benchmark our room labeling model separately, we compare it against different strategies within the domain. As the previous works used ground-truth object-to-room association, all baselines in this study are given the same privilege to ensure a fair comparison.
\begin{enumerate}
    \item \textit{Statistical Baseline:} This model estimates room labels by calculating the product of conditional probabilities for each object present, using ground truth object co-occurrence data~\cite{chen2023leveraging}.
    \item \textit{Zero-Shot LLM Baseline:} Utilizes ~\cite{black2021gpt} to generate descriptive strings about room compositions and applies a scoring method to identify the most appropriate room label~\cite{chen2023leveraging}.
    \item \textit{Embedding-Based Fine-Tuning:} We use RoBERTa~\cite{liu2019roberta} to create embeddings from strings describing objects in a room. These embeddings are then used to predict room types through a multi-layer perceptron~\cite{chen2023leveraging}.
    \item \textit{GNN Baseline:} Objects within a room are encoded as a graph, with nodes connected based on the proximity of objects. The graph is passed through SAGEConv~\cite{Hamilton2017InductiveRL} layers to predict room label logits, which are aggregated to determine the final room classification~\cite{gulshan2021gcexp}.
\end{enumerate}

\textbf{Ablation Study.} To assess the impact of using a self-attention transformer to find the room-label aligned embedding, we do an ablation study, where the room embedding is found by averaging the object CLIP embeddings for all the objects present in that room. In this study, we assign the room label by calculating the cosine similarity of the averaged embedding with the CLIP text embeddings of room phrases. The results are included in Table~\ref{tab:label_performance} in the third last row.
Additionally, we test a variant of our method where we add a simple single-layer perceptron to our model's architecture to predict logits for the room labels directly from the output \(e_{CLS}\) token embedding, trained using cross-entropy loss. The results for this are included in Table~\ref{tab:label_performance} in the second last row.

\begin{table}[htbp]
    \centering
    \caption{\textbf{Comparison of Room Labeling Models.}}
    \begin{tabular}{lcccc}
        \toprule
        \textbf{Method} & \textbf{Precision} & \textbf{Recall} & \textbf{F1} & \textbf{mAP} \\
        \midrule
        Statistical~\cite{chen2023leveraging} & 49.49 & 50.50 & 42.48 & 28.35 \\
        Zero Shot LLM~\cite{chen2023leveraging} & 46.69 & 45.34 & 43.00 & 48.07 \\
        Embedding-Based~\cite{chen2023leveraging} & 60.07 & 66.40 & 62.75 & 67.17 \\
        GNN Model~\cite{gulshan2021gcexp} & 58.79 & 65.59 & 61.38 & 67.45 \\
        \midrule
        Averaged Object Embeddings & 66.28 & 60.98 & 60.94 & 69.02 \\
        Ours (Logits) & 75.31 & 75.61 & 74.63 & 79.04 \\
        Ours (Contrastive) & \textbf{77.97} & \textbf{76.83} & \textbf{75.43} & \textbf{79.12} \\
        \bottomrule
    \end{tabular}
    \label{tab:label_performance}
    \vspace{-10pt} %
\end{table}

\textbf{Results.} As shown in Table~\ref{tab:label_performance}, our proposed CLIP Self-Attention Transformer model outperforms the next best algorithm, the embedding-based fine-tuning approach, by 10-12\% in both F1 and mAP metrics. This improvement is significant, especially for room types with fewer examples, addressing the challenge of long-tail distribution.  This specifically highlights the utility of our model in finding the room label-aligned embedding. 

\subsection{Complete Pipeline} \label{subsec:expt-complete-pipeline}
We evaluate our end-to-end pipeline on the Matterport3D dataset. We show the results for room segmentation and labeling in Table~\ref{tab:complete_pipeline} and compare it against the rich semantic results from RoomFormer~\cite{yue2023connecting}, the current SoTA in room segmentation and labeling. To demonstrate the difference between the output of our method and other open-vocabulary 3D scene-understanding methods, we show a qualitative comparison in Fig.~\ref{fig:qualitative-wide}. The comparison with OpenScene~\cite{peng2023openscene} shows that as the model computes per-point features, it is unable to correctly highlight the dense 3D region for a specific room type. We also contrast our results with the output from the 3D instance segmentation method OpenMask3D~\cite{takmaz2023openmask3d}, to show that it cannot highlight dense regions given a room type. Although it can correctly identify objects for simpler queries like \texttt{bedroom} and \texttt{kitchen}, it struggles with concepts like \texttt{living room} and \texttt{study room}. These results clearly highlight the difference in the room-level understanding of 3D scenes between previously published methods and our proposed pipeline. Moreover, our room and transition region outputs and room label outputs, enable the downstream applications such as natural-language laden instructions for navigation.

\begin{table}
    \centering
    \caption{\textbf{Complete Pipeline Evaluation.}}
    \begin{tabular}{cc}
        \toprule
         \textbf{Method}& \textbf{mAP @ 0.5 IoU}\\
         \midrule
         RoomFormer~\cite{yue2023connecting}& 23.52\\
         GT Room Masks + Our Room labeling & 79.12\\
         Ours (Complete Pipeline)& 74.02\\
         \bottomrule
    \end{tabular}
    \label{tab:complete_pipeline}
    \vspace{-10pt} %
\end{table}

\section{Limitations and Conclusion}

In cases where the room was not visible from a top-down view (due to noise or due to the room being very small),  the room segmentation pipeline was unable to recognise it. In cases where there were no objects identified in a room by the instance mapping pipeline, the labeling model was unable to generate a CLIP vector for the room. Some environments such as open plan houses are better organised into ``zones'' rather than ``rooms''. Since we rely on walls to segment, we cannot identify sub-room ``zones''. 

To summarise, our model extracts a natural-language queryable semantic topological map of 3D indoor scenes. We utilise a 3 channel Mask RCNN-based segmentation model to identify rooms and transition regions. We then utilise a self-attention transformer to generate room-label aligned CLIP features for our rooms based on their contained objects. We outperform the SoTA floorplan segmentation and labeling models on two challenging datasets. Additionally, we enable open-vocabulary room-level natural language queries on the scene. Future work may explore methods to segment a scene into zones. Language-queryable topological segmentation of a 3D scene into rooms can enable a robot to better reason about localization and enhance downstream tasks like planning~\cite{rana2023sayplan} and navigation~\cite{garg2023robohop}; integration of our method in such pipelines will be an exciting future work.

{\small
\bibliographystyle{ieee}
\bibliography{egbib}
}

\addtolength{\textheight}{-12cm}   %

\end{document}